\documentclass{article}

\usepackage{arxiv}
\usepackage[utf8]{inputenc}
\usepackage[T1]{fontenc}
\usepackage{hyperref}
\usepackage{url}
\usepackage{booktabs}
\usepackage{amsfonts}
\usepackage{amsmath}
\usepackage{graphicx}
\usepackage{microtype}
\usepackage{natbib}
\usepackage{xcolor}
\usepackage{enumitem}

\title{Verifiable Self-Evolution for Open-Ended Dialogue Skills via Future-Feedback Prediction}

\author{
  ChaoJin Zhao \\
  ByteDance \\
  \texttt{zhaochaojin@bytedance.com}
  \And
  Xuan Jiang \\
  ByteDance \\
  \texttt{jiangxuan.1217@bytedance.com}
}

\date{}
\renewcommand{\shorttitle}{Verifiable Dialogue Skill Evolution}
\hypersetup{
  pdftitle={Verifiable Self-Evolution for Open-Ended Dialogue Skills via Future-Feedback Prediction},
  pdfauthor={ChaoJin Zhao, Xuan Jiang},
  pdfkeywords={agent skills, self-evolution, dialogue evaluation, user feedback prediction}
}

\newcommand{\ctx}{\mathcal{C}}
\newcommand{\hist}{\mathcal{H}}

\newcommand{\fskill}{\mathcal{S}_{F}}
\newcommand{\askill}{\mathcal{S}_{A}}

\begin{document}
\maketitle

\begin{abstract}
Textual skills provide a lightweight way to improve frozen language-model agents, but their self-evolution normally requires a stable validation signal. Such signals are natural in mathematics or code, where an answer can be checked after it changes, yet are problematic in open-ended dialogue: changing the assistant response also changes the user's next reaction, so a logged reaction cannot directly evaluate a counterfactual response. We propose \emph{future-feedback skill evolution}, which first redirects self-evolution from prescribing the current answer to predicting whether the observed answer will lead to a positive or negative subsequent user signal. This prediction task is verifiable on fixed logged tuples and therefore supports validation-gated textual optimization. The evolved feedback skill captures interpretable criteria for response quality and can subsequently serve as a diagnostic and optimization target for answer skills. On a proprietary, privacy-preserving sales-assistant dataset, careful quality filtering and a balanced resolved/unresolved split yield more than 75\% prediction accuracy. Beyond this result, the central contribution is a formulation that converts otherwise moving conversational feedback into a fixed offline learning target, enabling reproducible skill evolution without placing every candidate skill in live traffic. We discuss the boundary between observational verification and counterfactual validity, and position the method as an offline optimization stage rather than a replacement for final human or online evaluation.
\end{abstract}


\section{Introduction}
Large language model (LLM) agents are increasingly controlled by textual \emph{skills}: persistent instructions that encode domain procedures, quality criteria, and tool-use policies. Compared with parameter training, skill editing is inexpensive, inspectable, and deployable without changing the underlying model. Recent work such as SkillOpt formalizes this idea as text-space optimization: an optimizer proposes bounded edits to a skill, and an edit is retained only when it improves a held-out score \citep{yang2026skillopt}.

The validation score is the critical assumption. In verifiable domains, an altered response can still be checked against a stable target. A mathematical proof can be re-evaluated, code can be executed, and structured outputs can be compared with known answers \citep{gao2023pal}. Open-ended dialogue lacks this invariance. Consider a logged interaction
\[
  x=(\ctx,\hist,Q_1,A,Q_2),
\]
where $\ctx$ is contextual and user information, $\hist$ is dialogue history, $Q_1$ is the current request, $A$ is the assistant response, and $Q_2$ is the user's subsequent utterance or behavioral feedback. If an answer skill changes $A$ into $A'$, the original $Q_2$ is no longer the reaction that would necessarily follow. Consequently, scoring $A'$ against the logged $Q_2$ conflates factual observation with an unobserved counterfactual.

This creates a practical obstacle for dialogue skill evolution. Without a fixed offline objective, each candidate answer skill would ideally require human judgment or online A/B traffic. Iterative deployment is slow, expensive, and risky; moreover, it provides no clean validation gate indicating whether textual self-evolution has converged.

We introduce a change of learning target. Instead of first evolving a skill that tells the assistant \emph{how to answer}, we evolve a feedback-prediction skill that estimates \emph{how a user will respond to an already observed answer}. Given $(\ctx,\hist,Q_1,A)$, the predictor estimates whether the next signal indicates resolved/accepted or unresolved/rejected. Unlike counterfactual answer evaluation, this task can be measured against a fixed logged label. Failed predictions are summarized into bounded additions, deletions, or replacements in the feedback skill; candidates are accepted only when they improve held-out performance.

The resulting feedback skill is useful beyond classification. To predict whether a user will accept an answer, it must encode operational distinctions between superficially plausible responses and genuinely resolving ones: tool success is not user adoption, a weak acknowledgment is not acceptance, a generic suggestion is not an actionable plan, and a response must cover the user's independent decision points. These rationales expose concrete directions for improving the answer skill.

Our contributions are:
\begin{itemize}[leftmargin=*]
  \item We formulate the moving-target problem that prevents direct reuse of logged next-turn feedback for offline evolution of open-ended answer skills.
  \item We propose validation-gated future-feedback skill evolution, turning logged human and behavioral signals into a fixed, verifiable objective for textual skill optimization.
  \item We identify a feedback--generation duality: the interpretable rules required to predict dissatisfaction also provide actionable diagnostics for answer generation.
  \item We report an industrial case study in which cleaned high-quality data and a balanced resolved/unresolved evaluation produce over 75\% accuracy, while explicitly separating observational prediction accuracy from counterfactual guarantees.
\end{itemize}

\begin{figure*}[t]
\centering
\includegraphics[width=\textwidth]{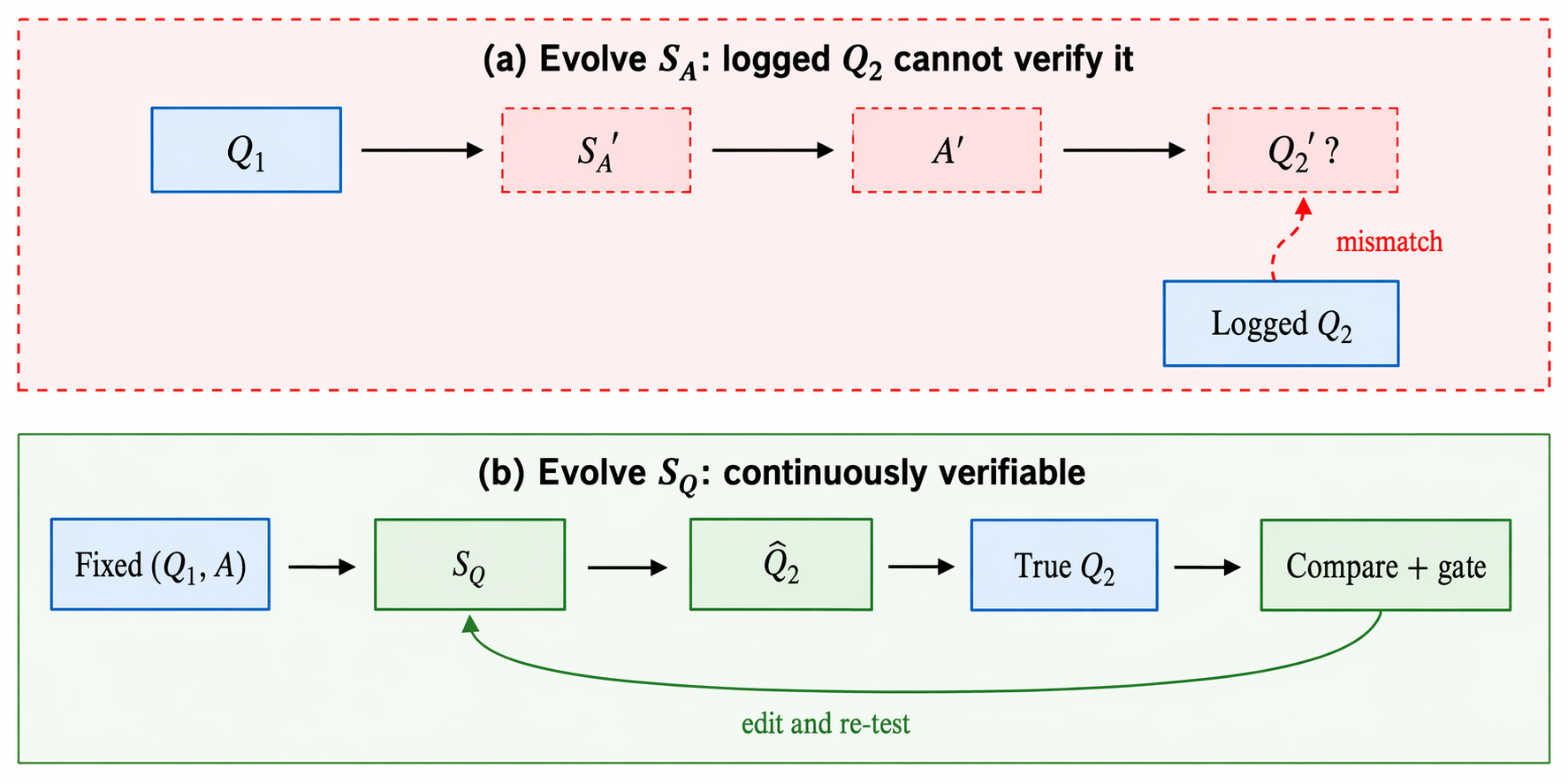}
\caption{Verification differs for the two skills. Revising $S_A$ produces a new answer $A'$ and an unknown subsequent signal $Q_2'$, so logged $Q_2$ cannot validate the candidate. Revising $S_Q$ keeps $(Q_1,A,Q_2)$ fixed, allowing every prediction to be compared with observed $Q_2$.}
\label{fig:concept}
\end{figure*}

Figure~\ref{fig:concept} makes the claim boundary explicit. Held-out logged data verifies evolution of the feedback-prediction skill itself. Its learned criteria can guide answer-skill edits, but the resulting counterfactual answers require fresh human judgments or online interactions; the original logged reaction is not a valid label for them.

\section{Related Work}
\paragraph{Textual skill and prompt optimization.}
Textual artifacts can serve as external, editable state for frozen models. SkillOpt applies optimizer-style discipline to this space through bounded edits, held-out validation, and rejection of non-improving candidates \citep{yang2026skillopt}. Our work adopts this validation-gated perspective but addresses a domain in which the validation target itself changes when the generated response changes.

\paragraph{User satisfaction estimation.}
User satisfaction has long been studied as an automatic evaluation signal for dialogue systems. The USS benchmark provides turn- and dialogue-level satisfaction annotations across multiple domains \citep{sun2021uss}. SPUR uses supervised iterative prompting to learn interpretable satisfaction rubrics from labeled examples \citep{lin2024spur}. These studies establish that satisfaction-related patterns can be learned from dialogue. Our focus is not merely estimating satisfaction; it is using a verifiable feedback-prediction task as the optimization substrate for self-evolving textual skills.

\paragraph{Predicting future dissatisfaction.}
\citet{see2021dissatisfaction} predict whether a user will express dissatisfaction in the next turn and use that signal to rank chatbot responses. This is closely related to our predictive target. We differ in treating the predictor's textual policy itself as the object of validation-gated evolution and in extracting its learned failure rationales as reusable skill knowledge.

\paragraph{Learned rewards and their limits.}
Human-feedback learning uses learned preference or reward models to optimize model behavior \citep{ouyang2022training}. Such models may fail under distribution shift or be exploited by optimization. Our proposal is deliberately narrower: the logged-data objective verifies the evolution of the feedback skill on observed interactions. It does not, by itself, prove that arbitrary answers optimized against that skill will satisfy users. Final deployment still requires human or online confirmation.

\section{Problem Formulation}
\subsection{Logged conversational feedback}
Let
\[
D=\{(X_i,A_i,Y_i)\}_{i=1}^{N}, \quad X_i=(\ctx_i,\hist_i,Q_{1,i}),
\]
where $Y_i\in\{0,1\}$ is derived from the subsequent user utterance or a behavioral signal. In our application, $Y=1$ means adopted or resolved and $Y=0$ means unresolved.

An answer skill $\askill$ induces a response distribution
\[
A\sim p_{\theta}(A\mid X,\askill),
\]
for a frozen target model with parameters $\theta$. The true utility of a new answer depends on an unknown reaction:
\[
U(A')=\mathbb{E}[Y'\mid X,A'].
\]
For a logged record containing $A$, only $Y$ is observed. Replacing $A$ with $A'$ does not preserve the label because generally
\[
p(Y\mid X,A) \neq p(Y\mid X,A').
\]
Thus, a fixed dataset cannot directly provide an unbiased validation score for arbitrary revisions of $\askill$.

\subsection{A verifiable surrogate task}
We instead optimize a feedback skill $\fskill$ that predicts the observed label:
\[
\hat{Y}=f_{\theta}(X,A;\fskill).
\]
For every candidate $\fskill'$, the held-out score
\[
J(\fskill')=\frac{1}{|D_{\mathrm{val}}|}
\sum_{(X,A,Y)\in D_{\mathrm{val}}}
\mathbb{1}[f_{\theta}(X,A;\fskill')=Y]
\]
is measurable without changing the interaction. Other deployment-relevant metrics, such as asymmetric costs for false positives, may be included in $J$.

This shift does not make the feedback model a perfect counterfactual oracle. It does make \emph{its own evolution} reproducible and rejectable on fixed data, which is the property needed for disciplined textual optimization.

\section{Method}
\subsection{Overview}
\begin{figure*}[t]
\centering
\includegraphics[width=\textwidth]{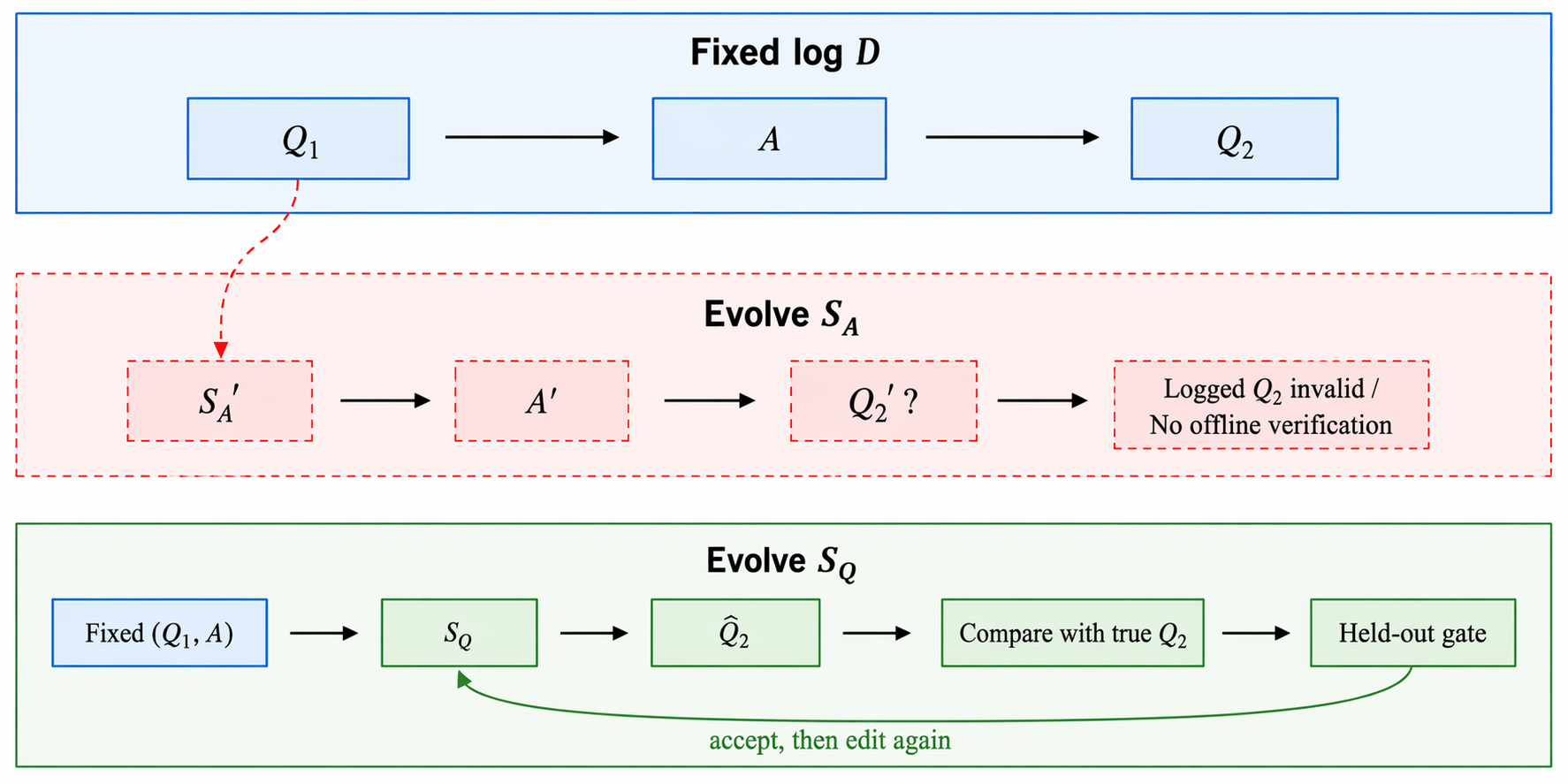}
\caption{Formal comparison with a fixed log. Evolving $S_A$ changes $A$ and leaves the subsequent $Q_2'$ unknown. Evolving $S_Q$ predicts the recorded $Q_2$, so every edit remains testable on the same held-out data.}
\label{fig:overview}
\end{figure*}

Figure~\ref{fig:overview} contrasts direct answer-skill evolution with our approach. Direct evolution tries to reuse a reaction attached to the old answer. Future-feedback evolution holds the logged answer fixed, learns a validated prediction skill, and then uses its interpretable criteria to diagnose answer quality.

\subsection{Feedback labels and data construction}
Signals may originate from explicit user feedback, textual acceptance or rejection, and product-side behavioral events. The implementation maps available evidence into a binary resolved/unresolved label. Silent termination is excluded when its meaning is ambiguous; this avoids noisy labels but introduces selection bias, discussed in Section~\ref{sec:limitations}.

Data quality is essential. Raw production logs contain incomplete turns, proactive outreach that is not a response to a user problem, duplicated or malformed records, weakly grounded labels, and missing conversational context. We apply quality filters so that each retained example has a clear turn boundary, a usable response, and a high-confidence subsequent label. We then construct train and test partitions with an approximately $1{:}1$ resolved/unresolved ratio. This prevents majority-class accuracy from obscuring whether the skill learns both outcomes.

\subsection{Validation-gated skill evolution}
Starting from an initial feedback skill $\fskill^{(0)}$, the predictor produces a label and rationale for each training example. At iteration $t$, an optimizer model receives high-value failures---especially confident false positives---and proposes a bounded textual edit $\Delta^{(t)}$:
\[
\tilde{\fskill}^{(t+1)}=\operatorname{Edit}(\fskill^{(t)},\Delta^{(t)}).
\]
Both the current and candidate skills are evaluated on the held-out set. The candidate is accepted only if it improves the primary validation objective; otherwise it is rejected:
\[
\fskill^{(t+1)}=
\begin{cases}
\tilde{\fskill}^{(t+1)}, & J(\tilde{\fskill}^{(t+1)})>J(\fskill^{(t)}),\\
\fskill^{(t)}, & \text{otherwise.}
\end{cases}
\]
A lexicographic variant can retain equal-accuracy candidates only when an asymmetric business utility improves. Penalizing false positives more strongly is useful when declaring a user satisfied despite an unresolved issue is costlier than conservatively predicting unresolved.

\subsection{From prediction rationales to answer guidance}
The evolved skill learns reusable distinctions. Representative rules include:
\begin{itemize}[leftmargin=*]
  \item successful tool execution or message delivery does not imply that the user adopted the result;
  \item greetings, product recommendations, reports, and calls to action do not by themselves resolve a user need;
  \item weak acknowledgments such as ``OK'' may indicate receipt rather than acceptance;
  \item parameter advice should specify an action and at least one operational anchor, such as a value, interface, time window, or re-evaluation condition;
  \item the response should cover the independent decision points in the user's latest request.
\end{itemize}

These rules can be transformed into diagnostics for $\askill$. For example, a response recommending that a user ``observe for several days and adjust the parameter appropriately'' appears helpful but lacks a target value and a trigger for further adjustment. The feedback skill predicts unresolved and explains the missing actionability; the explanation can then be distilled into an answer-skill instruction. In this paper, we establish this bridge conceptually and through observed learned rules, but do not claim a separately measured end-to-end improvement of $\askill$.

\section{Industrial Case Study}
\subsection{Setting}
We evaluate the method on privacy-preserving interactions from a production sales-assistant scenario. Inputs include available context, dialogue history, tool-use summaries, and the assistant's current response. Labels indicate whether subsequent user or behavioral evidence supports resolved/adopted versus unresolved/not adopted. All examples and operational identifiers are anonymized, and no raw dialogue is released.

The latest evaluation uses a curated high-quality subset rather than the earlier unfiltered extraction. Train and test sets are precisely partitioned and balanced to approximately equal positive and negative proportions. Because this paper emphasizes the formulation and the data are proprietary, we report the verified aggregate result without disclosing sensitive counts or traffic statistics.

\subsection{Result}
On the balanced held-out set, the evolved feedback skill achieves \textbf{more than 75\% accuracy}. A random or constant classifier on this construction has approximately 50\% accuracy, so the result indicates that the skill extracts meaningful interaction-level signals rather than exploiting label prevalence.

Two engineering choices were decisive. First, data cleaning removed low-quality and weakly labeled interactions whose conversational relation was ambiguous. Second, precise partitioning produced a balanced test set, making accuracy directly interpretable across resolved and unresolved cases. We treat these as part of task construction rather than as a novel modeling contribution: a verifiable objective is useful only when the labels and partitions faithfully represent the behavior to be predicted.

The learned rules also provide qualitative evidence of convergence toward user-centered criteria. The predictor becomes less likely to equate fluent sales language, completed workflows, or successful tools with actual resolution, and more likely to require coverage, specificity, and an actionable next step.

\subsection{What the result establishes}
The experiment supports three claims. First, next-feedback prediction is learnable from cleaned conversational records. Second, textual skill edits can be evaluated and gated on fixed held-out data. Third, the evolved artifact is interpretable enough to expose recurring answer defects.

The result does \emph{not} establish that 75\% observational accuracy is sufficient to optimize arbitrary counterfactual answers safely. When the answer distribution moves, the predictor may also shift in accuracy. Our method therefore reduces the need for repeated online testing during skill search, but does not eliminate a final human preference study or controlled online validation for a deployed answer skill.

\section{Discussion}
\subsection{Why predict feedback instead of directly judging answers?}
A direct LLM judge can assign a score to a new answer, but its score may encode generic preferences rather than the target users' actual behavior. Future-feedback prediction ties the skill to observed human or behavioral outcomes. It also produces an ordinary supervised validation loop: every prediction can be compared with a fixed label, and every textual edit can be accepted or rejected reproducibly.

\subsection{The feedback--generation duality}
Prediction and generation are not identical, but they share a quality model. A predictor cannot reliably forecast dissatisfaction without representing omissions, ambiguity, non-actionability, unsupported claims, and mismatches with user intent. This creates a useful duality:
\[
\text{predictive failure criteria}
\quad\Longrightarrow\quad
\text{candidate generation constraints}.
\]
The feedback skill is therefore both an evaluator and a repository of response-quality knowledge. Future work can automate the compilation of its rationales into answer-skill edits and validate those edits with a one-time counterfactual human study.

\subsection{Operational value}
The framework changes the role of online evaluation. Instead of exposing every candidate skill to users, developers can perform most iterations offline, reject non-improving candidates, inspect learned rules, and deploy only the final candidate for controlled confirmation. This is especially valuable in customer-facing systems where careless exploration can degrade trust.

\section{Limitations}
\label{sec:limitations}
\paragraph{Counterfactual distribution shift.}
The predictor is validated on logged answers. Optimizing an answer skill may generate out-of-distribution responses that exploit predictor weaknesses. The feedback skill should therefore be viewed as a validated observational model and diagnostic oracle, not a proof of human satisfaction for arbitrary $A'$.

\paragraph{Silent users.}
Interactions with no interpretable subsequent signal are excluded. Some may represent silent dissatisfaction, so the resulting labels cover explicit or behaviorally observable resolution more reliably than total user satisfaction.

\paragraph{Data curation and reproducibility.}
The strongest reported result depends on proprietary cleaning rules and a private production dataset. This is common in industrial dialogue research but limits external reproduction. Public satisfaction datasets such as USS could test cross-domain generality in future work \citep{sun2021uss}.

\paragraph{Binary feedback.}
Resolved/unresolved labels simplify graded, personalized satisfaction. A user can accept an answer while remaining partly dissatisfied, or reject a correct answer for reasons outside response quality.

\paragraph{No additional answer-skill experiment.}
Consistent with the scope of this study, we do not report a new human evaluation or online A/B test of an answer skill distilled from the feedback predictor. We therefore present rationale-to-guidance transfer as a supported mechanism and research direction, not as a completed causal result.

\section{Conclusion}
Open-ended dialogue breaks the fixed-target assumption underlying validation-gated skill evolution: changing an answer changes the feedback that would follow. We proposed future-feedback skill evolution, which makes the first optimization target the prediction of logged user feedback rather than the direct generation of a counterfactual answer. This yields a fixed offline objective for evolving an interpretable feedback skill. In a curated and balanced industrial dataset, the method achieves over 75\% held-out accuracy and learns criteria that distinguish superficial completion from genuine resolution. The broader lesson is that verifiability in open dialogue can be recovered by choosing a prediction target whose outcome has already been observed. This does not remove the need for final human validation, but it makes the path to that validation more controlled, inspectable, and reproducible.

\bibliographystyle{unsrtnat}
\bibliography{references}

\end{document}